\documentclass[runningheads]{llncs}
\usepackage{epsfig}
\usepackage{subcaption}
\usepackage{calc}
\usepackage{amssymb}
\usepackage{amstext}
\usepackage{amsmath}
\usepackage{multicol}
\usepackage{multirow}
\usepackage{pslatex}
\usepackage[bottom]{footmisc}

\usepackage{tablefootnote}
\usepackage{booktabs}
\usepackage{hyperref}
\usepackage{placeins}
\usepackage{newfloat}
\usepackage{algorithm}
\usepackage{algorithmicx,algpseudocode,placeins}

\DeclareMathOperator*{\argmin}{arg\,min} 

\newcommand{\norm}[1]{\left\lVert#1\right\rVert}
\newcommand*{\placeholderfunc}{\makebox[1ex]{\textbf{$\cdot$}}}%

\usepackage[T1]{fontenc}
\usepackage{graphicx}
\usepackage{color}

\makeatletter
\newlength{\phaserulewidth}
\newcounter{phase}[algorithm]
\newcommand{\setphaserulewidth}{\setlength{\phaserulewidth}}
\newcommand{\phase}[1]{%
  \vspace{-1.25ex}
  \Statex\leavevmode\llap{\rule{\dimexpr\labelwidth+\labelsep}{\phaserulewidth}}\rule{\linewidth}{\phaserulewidth}
  \Statex\strut\refstepcounter{phase}\textit{Part~\thephase~--~#1}
  \vspace{-1.25ex}\Statex\leavevmode\llap{\rule{\dimexpr\labelwidth+\labelsep}{\phaserulewidth}}\rule{\linewidth}{\phaserulewidth}}
\makeatother
\setphaserulewidth{.7pt}

\begin{document}
\newcommand{\algoname}{\textit{FedAcross+ }}

\title{Domain Borders Are There to Be Crossed With Federated
Few-Shot Adaptation}
\titlerunning{FedAcross+}
%
%
\author{Manuel Röder\inst{1,2,3}\orcidID{0009-0003-4907-3999} \and
Christoph Raab\inst{4}\orcidID{0000-0002-6988-353X}
\and
Frank-Michael Schleif\inst{1}\orcidID{0000-0002-7539-1283}}
\authorrunning{M. Röder et al.}
%
\institute{Faculty of Computer Science and Business Information Systems, Technical University of Applied Sciences Würzburg-Schweinfurt, Würzburg, Germany\\
\email{\{manuel.roeder, frank-michael.schleif\}@thws.de}
\and
Faculty of Technology, Bielefeld University, Bielefeld, Germany
\and
Center for Artificial Intelligence, Würzburg, Germany\\
\and
IAV GmbH, Berlin, Germany\\
\email{christoph.raab@iav.de}}
\maketitle              
\begin{abstract}
Federated Learning has emerged as a leading paradigm for decentralized, privacy-preserving learning, particularly relevant in the era of interconnected edge devices equipped with sensors.
However, the practical implementation of Federated Learning faces three primary challenges: the need for human involvement in costly data labelling processes for target adaptation, covariate shift in client device data collection due to environmental factors affecting sensors, leading to discrepancies between source and target samples, and the impracticality of continuous or regular model updates in resource-constrained environments due to limited data transmission capabilities and technical constraints on channel availability and energy efficiency.
To tackle these issues, we expand upon an efficient and scalable Federated Learning framework tailored for real-world client adaptation in industrial settings.
This framework leverages a pre-trained source model comprising a deep backbone, an adaptation module, and a classifier running on a powerful server.
By freezing the backbone and classifier during client adaptation on resource-constrained devices, we allow the domain adaptive linear layer to handle target domain adaptation, thus minimizing overall computational overhead.
Furthermore, this setup, designated as \textit{FedAcross+}, is extended to encompass the processing of streaming data, thereby rendering the solution suitable for non-stationary environments. 
Extensive experimental results demonstrate the effectiveness of \emph{FedAcross+} in achieving competitive adaptation on low-end client devices with limited target samples, successfully addressing the challenge of domain shift.
Moreover, our framework accommodates sporadic model updates within resource-constrained environments, ensuring practical and seamless deployment.

\end{abstract}

\section{Introduction}
\label{sec_intro}
Traditional machine learning relies on a centralized data center to store and aggregate training data collected from local devices such as mobile phones, drones, or thin clients.
This method has proven impractical for real-world applications, as it demands substantial effort to gather and label data from various sources while complying with data protection regulation.

In ~\cite{pmlr-v54-mcmahan17a}, \emph{Federated Learning} (FL) was introduced to mitigate the security risks and costs associated with traditional model implementations. This architecture allows multiple edge devices to collaboratively learn a global machine learning model managed by a central server, while the local data remains on the client devices.
The recent epoch has witnessed remarkable advancements in the realms of hardware and software technologies.
A salient trend that has emerged is the ubiquitous adoption and interconnectivity of sensor-equipped devices at the edge~\cite{10.1145/3478680}, often employed in industrial production settings.
These technological progressions have unveiled novel prospects for data acquisition, analysis, and automation in manufacturing environments.
This trend, combined with the growing adoption of 5G-enabled edge devices, has significantly boosted the appeal of FL for real-world industry applications and research purposes~\cite{Hard2018FederatedLF,yang2018applied,10.1145/3298981,Yang2020FederatedML}.

\begin{figure*}
    \includegraphics[width=\textwidth]{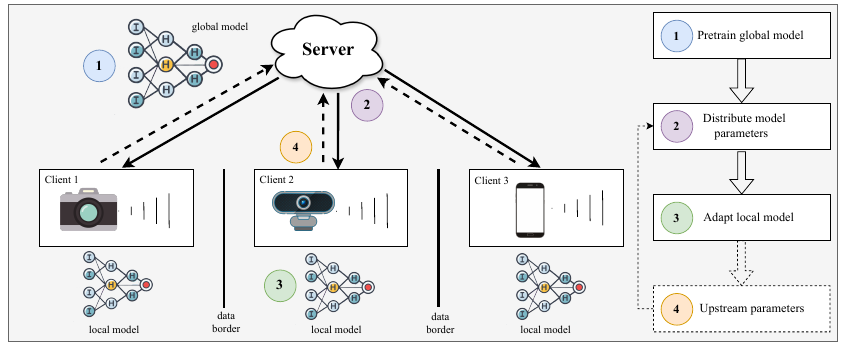}
    \caption{Featured approach overview for federated client adaptation. \\Adapted from~\cite{roder2024orig}.}
    \label{fig:teaser}
\end{figure*}

A prevalent real-world application where the conventional FL approach is not directly suitable involves the classification of waste items~\cite{WeSort_AI,DBLP:journals/corr/abs-2303-14828} by resource-constrained client devices outfitted with visual sensors situated at various waste sorting facilities (refer to the conceptual design overview depicted in Figure~\ref{fig:teaser}).
In an era of continuously escalating quantities of refuse, waste sorting poses a critical challenge for numerous communities, and intelligent sensor systems constitute a pivotal component in the diverse strategies aimed at addressing this challenge~\cite{doi:10.1021/acssuschemeng.1c05013}. Within this scenario, each client's local model undergoes training utilizing isolated processing units, while data generation occurs locally and remains decentralized.

\emph{Cross-device federated learning}~\cite{10.1561/2200000083} addresses the previously mentioned obstacles of collaborative learning across numerous devices with constrained data accumulation, costly labelling, and restricted sharing.

\begin{figure*}
    \centering
    \includegraphics[width=0.85\textwidth]{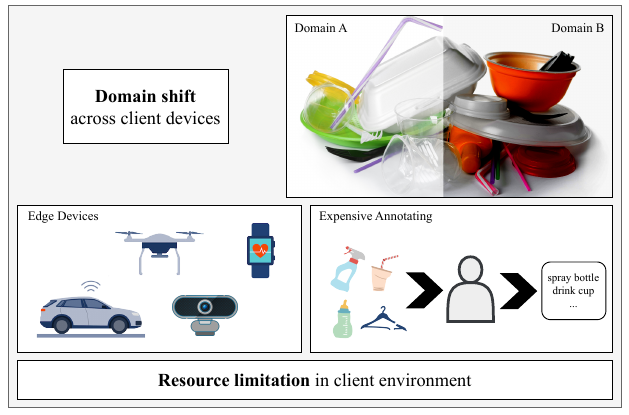}
    \caption{Challenges of real-world FL applications: Domain shift across clients and resource limitations (data sparsity, label scarcity, storage restrictions) on client edge devices. Adapted from~\cite{roder2024orig}.}
    \label{fig:chall}
\end{figure*}

Additionally, there are numerous external factors that significantly impact both the availability and the quality of sampled sensor data.
For instance, in the case of visual sensors, domain shifts can occur in the images captured by these sensors (refer to Figure~\ref{fig:chall}).
Such shifts may arise from variations in lighting conditions or changes in the surrounding environment, which subsequently influence several image attributes. These attributes include, but are not limited to, brightness, contrast, color temperature, perspective, and the level of noise present in the images~\cite{wilds2021}.
In general terms, the complex task of adjusting a system that has been initially trained within a source domain to enhance its performance within a target domain is identified as \emph{source-to-target adaptation}~\cite{DBLP:journals/ijon/RaabRS22}.
Given the constraints on hardware resources available to clients, including computing power, transmission capacity, and memory usage, it is imperative to develop a training algorithm that reduces the load on the client-side. This algorithm must achieve this reduction without compromising the accuracy and efficacy of the model training process.
As a result, this research concentrates on incorporating a well-established pre-training and fine-tuning strategy within the framework of FL.
The primary objective is to convert domain-specific features into a task-invariant metric space, thereby alleviating the impact of domain shifts while operating under stringent resource constraints.
To address transmission costs and privacy restrictions (see Section \ref{client_share_sec} for details) in cross-device FL, we introduce an innovative solution based on \emph{prototypes}~\cite{snell2017prototypical}, a concept previously underexplored in FL across diverse domains.
This method effectively reduces data transfer overhead and minimizes computational costs for inference on client devices.
Our approach involves computing local prototypes by averaging the feature vectors within each class, thereby achieving memory efficiency.
For client-side label prediction, the process entails comparing the distances between the projected input representations and the class prototypes within the embedding space. This distance-based comparison optimizes CPU usage, ensuring a lightweight and efficient inference process on client devices.
By exclusively exchanging prototypes, we significantly mitigate transmission costs and enhance the overall efficiency of the FL system.
As detailed below, this article introduces an extended FL framework, \textit{FedAcross+}, which addresses the real-world challenges of data sparsity on isolated clients and the domain shifts that occur across clients deployed in diverse industrial environments.

Previous work~\cite{roder2024orig} relies on few-shot support sets containing equally distributed, labeled class instances for fine-tuning and prototype generation.
While effective, this approach depends on static, pre-collected data batches, which can be inefficient and impractical in dynamic environments with continuously generated data points.
To overcome these limitations, we are extending the existing framework to incorporate \emph{sampling from data streams} in this work, enabling real-time model updates and more responsive learning.
\\
\noindent
\textbf{Main contributions:} 
\begin{enumerate}
    \item We present a computation-efficient FL approach designed to address target adaptation challenges arising from limited labeled samples and distributional shifts across siloed devices in real-world applications. This approach aims to enhance the performance of local models on individual devices for downstream classification tasks, while concurrently iterating on the optimization of global model parameters to facilitate the onboarding of new FL clients.
    \item We implement a ready-to-deploy, efficient, and highly scalable end-to-end FL solution, realized using PyTorch Lightning~\cite{Falcon_PyTorch_Lightning_2019} and Flower~\cite{beutel2020flower}, available on Github\footnote{\href{https://github.com/cairo-thws/FedAcrossPlus}{https://github.com/cairo-thws/FedAcrossPlus}}.
    \item Building on the discourse in~\cite{roder2024orig}, we extend the existing framework to facilitate sampling from distributed data streams.
    \item We conduct a comprehensive evaluation of our method in a waste item classification scenario using domain adaptation benchmark data sets that replicate production conditions.
    Our approach demonstrates competitive adaptation performance compared to state-of-the-art methods.
    This scenario is particularly challenging and encompasses a wide range of real-world complexities in cross-device FL.
    Importantly, our method is not restricted to waste item classification; it can be effectively applied to a variety of other use cases, including supply chain optimization, smart grid energy management, and epidemic and disease surveillance, emphazising its versatility and potential.
\end{enumerate}

\section{Related Work}
Previous studies have extensively investigated the limitations of device-based FL, where data remains isolated within individual entities or organizations.
Researchers have undertaken investigations to address the challenges associated with this siloed approach, including data privacy concerns, communication overhead, and model performance degradation~\cite{9599589}.
However, existing approaches still face limitations in efficiently utilizing information present in siloed data while maintaining privacy.
Achieving efficient information transfer and compact encoding is crucial to overcome the barriers of crossing device borders with minimal transmission costs and attaining robust generalization capabilities~\cite{9599589}.
This is getting even more challenging in realistic scenarios, where often only weakly labeled data points are available.

In centralized scenarios, techniques have been proposed using \emph{Few-shot learning} (FSL) to enable models to learn effectively from limited labeled data~\cite{article_wang_2020,10.1145/3582688,DBLP:conf/cvpr/Hu0SKH22}.
These approaches typically rely on fine-tuning strategies to enhance the model's ability to generalize and adapt to new tasks, even with only a few labeled training samples.
Two primary training paradigms have emerged: \emph{learning by transfer}~\cite{Dhillon2020A}, where a deep neural network is initially trained on a source data set and subsequently fine-tuned for a downstream few-shot learning task, and \emph{meta-learning} approaches as exemplified by~\cite{snell2017prototypical,vinyals_matching_2016}, where incremental parameter updates encode task-specific background information during the model optimization process.

Additionally, the siloed setting and common phenomena of error-prone measurement devices introduce various effects of cross-domain shifts.
Cross-domain learning specifically focuses on transferring knowledge from a source domain to a target domain, even when the data's characteristics or distribution significantly differ.
Extensive research in this area has explored \emph{domain adaptation} (DA) techniques, including domain alignment, feature mapping, and instance re-weighting, to enhance model performance when applied to unseen target domains~\cite{DBLP:journals/ijon/RaabRS22}.
Furthermore, \emph{test-time adaptation methods}~\cite{nado2021evaluating,zhang2022lccs} have demonstrated that even with a limited number of labeled data points per class from a target domain, it is possible to optimize domain-specific adaptation parameters effectively. These methods provide a robust framework that we can leverage to design and refine our model architecture, ensuring that it adapts efficiently to varying domain-specific characteristics.

\emph{Cross-domain few-shot learning} (CD-FSL) addresses a centralized approach to the aforementioned challenges, focusing on the effective and rapid learning of relevant information from only a few samples while simultaneously managing the distributional shift between source and target data.
Recent studies on CD-FSL~\cite{guo2020broader} have demonstrated that transfer learning approaches outperform state-of-the-art meta-learning methods on few-shot learning benchmarks across multiple domains.
Therefore, transfer learning offers a practical solution to avoid computationally intensive gradient update calculations for client models operating on edge devices.
It also enables the delegation of the substantial task of training a source model on a large data set to a well-equipped server instance.

The concept of \emph{prototypes} in the context of FL has already been examined from several perspectives.
FedProto~\cite{tan2021fedproto} has clients train local models to minimize classification error while keeping local prototypes close to the global aggregated prototypes.
FedPCL~\cite{fedpcl2022} uses prototype-wise contrastive learning where clients build personalized representations by contrasting against local and global prototypes.
FedNH~\cite{dai2023tackling} improves local models by enforcing uniformity and semantics in the class prototypes.
FedGPD~\cite{fedgpd2024} introduces a global prototype distillation loss to align local instances with the global class prototypes.
FedRFQ~\cite{fedrfq2024} employs a softpool mechanism to mitigate prototype redundancy and failure, and uses Byzantine fault tolerance for secure aggregation against poisoning attacks.
FPL~\cite{10203389} constructs unbiased prototypes as fair convergence targets and uses consistency regularization and 
FedTGP~\cite{zhang2024fedtgp} proposes trainable global prototypes with an adaptive margin loss.
However, these methods ignore communication costs or do not address client-side resource limitations and require a substantial amount of locally labeled target data.

Our strategy integrates ideas from various fields to address these issues and focuses on optimizing the local model of each client to adapt to its individual task to be solved. 
One particular variant of FL is commonly referred to as \emph{Personalized Federated Learning} (pFL)~\cite{Tan2021TowardsPF}, and it specifically aims to address the challenges posed by data heterogeneity across different clients.
Similar to the idea of our contribution, pFL allows each client to adapt the global model to their local data, often through additional local training steps or by incorporating personalized components into the downstream model.
Our approach shares many common elements with pFL, including the application of data augmentation techniques, adaptive model design and the use of transfer learning.

\emph{FL for data streams} and dynamic environments is a rapidly advancing field with significant research efforts aimed at addressing the unique challenges of continuous data generation and heterogeneous environments.
A general FL algorithm for data streams focuses on weighted empirical risk minimization to handle continuously generated data~\cite{marfoq2023federated}.
FedStream~\cite{mawuli2023fedstream} is another prototype-based framework designed for distributed concept-drifting data streams, employing error-driven representative learning and metric-learning-based prototype transformation.
Other notable contributions in this domain include the development of client clustering algorithms for drift adaptation~\cite{jothimurugesan2023federated}, which address the staggered nature of concept drifts across different clients.
These algorithms enhance the ability of FL systems to react to local drifts and maintain high accuracy.
In this work, we shift our focus from concept drift detection to enhancing the existing FedAcross framework without requiring a complete architectural overhaul. Accordingly, we incorporate a suitable sampling strategy designed to deliver data points in a format that aligns seamlessly with the input expectations of FedAcross as outlined in Section~\ref{client_sample_sec}.

\section{Methodology}
Build upon the concepts and assumptions of FedAcross~\cite{roder2024orig}, we outline the main prerequisites as well as fundamental theoretical background in Section~\ref{prerequisites_sec}.
Section~\ref{server_model_sec} details the end-to-end source model training procedure and Section~\ref{client_adapt_sec} explores the on-client model adaptation process under the constraints of minimal labeled sample availability and domain shift.
Subsequently, in Section~\ref{client_infer_sec}, we introduce a computation-efficient, prototype-based client inference pipeline.
Section~\ref{client_share_sec} discusses data privacy aspects and prototype upstreaming options.
Finally, Section~\ref{client_sample_sec} extends the existing static framework to support dynamic data streams and presents the pseudo-code for \textit{FedAcross+}.

\subsection{Prerequisites}\label{prerequisites_sec}
Our methodology considers a publicly accessible, labeled source data set $D_S =\{X_S,Y_S\} = \{x_l,y_l\}_{l=1}^N \stackrel{i.i.d.}{\sim} p_S(X_S,Y_S)$ in the source domain $S$ distributed by $p_S$.
For each federated client $i$ we suppose a target data set $D_{T_i} \stackrel{i.i.d.}{\sim} \{x_l,y_l\}_{l=1}^{k_i} \stackrel{i.i.d.}{\sim} p_{T_i}(X_{T_i},Y_{T_i})$ in the target domain $T_i$.
Without loss of generality, we assume that the number of labeled samples per class is $k = k_i = \{0, 3, 5, 10\}$ for client $i$.
Data samples are given as $x_l \in \mathbb{R}^d$, $d$ as number of features and $y_l \in Y$, $Y$ as a discrete label space, common for source and target domains, $|Y|=L$.
The observed distributions $p_S(X_S,Y_S) \neq p_{T_i}(X_{T_i},Y_{T_i})$ underlie domain shift.
Similar as $x_l$, we introduce class-wise prototypes, designated as $\omega^{(n)}$, for the respective source and target domains under inspection, where $n$ denotes the class index in $Y$.
A centralised server instance handles the resource-intensive part of model learning, while the training data is stored in separate data units on geographically distributed clients with limited communication and processing power.
The key goal is to train a classifier model that can generalize to related target domains (see "DomainA" and "DomainB" in Figure~\ref{fig:chall} as an example for source and target domain differences).
With $f(\phi)$ being a feature extractor parameterized by $\phi$ and $\mathcal{A}(\psi)$ an adaptation module parameterized by $\psi$.
Lastly, a classifier $g(\nu)$ is generated, parameterized by $\nu$.
The approach is applicable to a range of tasks, although it is assumed that the feature space is related to image processing.
Given a FL setup as outlined in Figure~\ref{fig:teaser}, a computationally powerful server can fully access the source domain data set $D_S$.
Additionally, each client $i$ participating in the distributed learning environment has exclusive access to its target domain data set $D_{T_i}$.
A comprehensive summary including the most important notation used in this work can be found in Table~\ref{notations_tab}.
\begin{table}[ht]
\caption{Notation Summary.\\Adapted from~\cite{roder2024orig}.}
\label{notations_tab}
\begin{center}

\begin{tabular}{ll}
\toprule
\text {Notation      } & \text {Description} \\
\hline
\text{$D_S$} & \text{source domain data set on server $S$}\\
\text{$D_{T_i}$} & \text{target domain data set on client $i$}\\
\text{$D^{spt}_{T_i}$} & \text{$k$-shot support set on client $i$, $D^{spt}_{T_i} \subset D_{T_i}$ }\\
\text{$N$} & \text{nr. of observed classes, $n = 1, ..., N$}\\
\text{$K$} & \text{nr. of labeled samples per class, $k = 0, ..., K$ }\\
\text{$f(\phi)$} & \text{feature extractor $f$ parameterized by $\phi$}\\
\text{$\mathcal{A}(\psi)$} & \text{adaptation module $\mathcal{A}$ parameterized by $\psi$}\\
\text{$\mu, \sigma$} & \text{batch normalization layer statistics}\\
\text{$\gamma, \beta$} & \text{batch normalization layer parameters}\\
\text{$g(\nu)$} & \text{classifier $g$ parameterized by $\nu$}\\
\text{$\omega^{(n)}_{i}$} & \text{prototype of class $n$ on client $i$}\\
\hline
\text{$U_{i}$} & \text{unlabeled stream of samples observed on client $i$}\\
\bottomrule
\end{tabular}
\end{center}
\bigskip
\end{table}

\noindent
In order to better reflect the real world conditions in the modelling process the following assumptions are made:
\begin{itemize}
     \item The number of annotated samples per class in the target domain data set $D_{T_i}$ is fixed by the value of $k$. The $k$-shot support set of client $i$ is then defined with $D^{spt}_{T_i} \subset D_{T_i}$, resulting in input-output pairs within each observed data set being equally distributed.
    In production environments, operators collect and annotate only $k$ samples for local model fine-tuning.
    \item The number of classes used for local fine-tuning can be limited to $n$ for each client individually, giving the client operator the option to separately select a subset of available classes to meet their needs.
\end{itemize}
The following paragraphs present a model architecture that adheres to the aforementioned constraints and addresses the challenges presented from beginning to end.

\subsection{Server Model Training}\label{server_model_sec}
\begin{figure}[ht]
\centering
	\includegraphics[width=\linewidth]{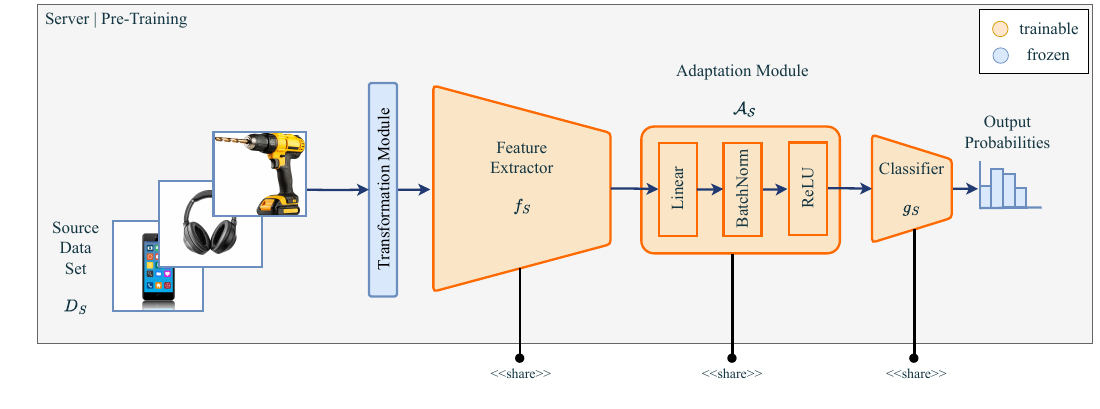}
	\caption{Model Architecture for server pre-training on source data set $D_S$. Module colors determine whether the parameters are frozen or trainable during training. Adapted from~\cite{roder2024orig}.
	  \label{fig:model_arch}
    }
\end{figure}

The proposed server model consists of four main components as visualized in Figure~\ref{fig:model_arch}.
This chapter provides a comprehensive elaboration of each of the aforementioned components, accompanied by a detailed justification of the respective design choice in light of the stated challenges.

The \emph{transformation module} $\mathcal{T}$ is the initial component of the server model. 
It maps the input $x_S$ taken from the source domain data set $D_S$ onto the output $\tilde{x}_S$ by applying a single augmentation chain that deploys concatenated affine transformations to alter model input data samples using $\tilde{x}_S \gets \mathcal{T}(x_S)$.
Affine transformations are a common and effective data augmentation technique for improving generalization and reducing overfitting in deep learning models for image classification tasks~\cite{perez2017effectiveness}.
These transformations, which include operations such as rotation, scaling, and translation, help create diverse synthetic training examples while preserving the essential content and labels of the original images.
By exposing models to a wider range of variations during training, affine data augmentation enhances their ability to learn robust features and generalize better to unseen data.
Further studies~\cite{kim2022finetune} have revealed that these data augmentation strategies offer advantages beyond just expanding data set size for CD-FSL.
These techniques have been shown to enhance the transfer learning process itself when applied to the target data set in downstream tasks.
The introduction of controlled variations in the limited target data, through the process of augmentation, enables the model to learn more robust and generalizable features, thus improving performance on image classification tasks in low data regimes.
Given that the server model is pre-trained on the entirety of the source data set $D_S$, we chose to implement the basic augmentation method for \emph{full fine-tuning} scenarios (whereby the all network parameters are updated) and integrate horizontal flipping, random resized cropping and color jittering into the pipeline of $\mathcal{T}$.

The \emph{feature extractor} $f$, parameterized by $\phi_S$, is the second model component that retrieves relevant features from the transformed input data $\tilde{x}_S$ and produces output data $\overset{\circ}{x}_S$, with $f(\tilde{x}_S) \mapsto \overset{\circ}{x}_S$, $\overset{\circ}{x}_S \in \mathbb{R}^m, m \ll d$.
In order to achieve an optimal compromise between the network depth required for the image classification problem and the constraint of maintaining the lowest possible run-time resource usage for client endpoints in the fine-tuning stage later on, a pre-trained ResNet-34~\cite{7780459} backbone with approximately $22$ million parameters is employed on the server side.

Subsequent to the feature extractor $f$, the next component of the source model is the \emph{adaptation module} $\mathcal{A}$ parameterized by $\psi_S$.
This constitutes the core of our contribution to this work, which is built upon the concept of \emph{task-specific adapters}~\cite{9879070} and \emph{universal templates for few-shot learning}~\cite{triantafillou2021learning}: a domain-adaptive linear layer allocates a dedicated set of conditional batch normalization parameters and linear layer weights for the source domain pre-training and each downstream target fine-tuning task.
This renders the adaptation module $\mathcal{A}$ as the sole entity responsible for \emph{Domain Adaptation} (DA)~\cite{DBLP:journals/corr/abs-1906-03950}. Formally, the server-side adaptation module is defined as
\begin{equation} \label{eq_adaptation_module}
    \begin{aligned}
        \mathcal{A}(\psi) & = \mathcal{A}(\mu, \sigma, \gamma, \beta, W, b; \overset{\circ}{x}_S ) \\
                          & = \frac{(W \overset{\circ}{x}_S + b) - \mu}{\sigma} \gamma + \beta
    \end{aligned}
\end{equation}
with $\overset{\circ}{x}_S$ denoting the output of the feature extractor, \{$\mu, \sigma$\} denoting the batch normalization layer statistics, \{$\gamma, \beta$\} are the batch normalization layer parameters and \{$W, b$\}, $W\in \mathbb{R}^{m \times m}, \, b\in \mathbb{R}$ are the weights and bias parameters of the linear layer, respectively. Parameter simplifications are possible by taking dependencies into account.
Explained in details, the feature extractor $f$ processes the input data and produces an output $\overset{\circ}{x}_S$, which serves as the input to the domain-adaptive linear layer.
The batch normalization step involves normalizing the output of the feature extractor using the mean ($\mu$) and variance ($\sigma$) computed from the batch statistics.
This normalization procedure helps in stabilizing the training process by ensuring that the input distribution to each layer remains consistent~\cite{lirevisitbatch2017}.
The parameters $\gamma$ and $\beta$ are used to scale and shift the normalized output as described in Equation~\ref{eq_adaptation_module}, respectively.
These parameters are specific to each domain, allowing the downstream model to adapt to different domain characteristics individually.
The normalized and scaled output is then passed through a linear transformation defined by the weights $W$ and bias $b$.
This transformation maps the input features to the desired output space.
Ultimately, the domain-adaptive linear layer offers several benefits for the fine-tuning phase of the client model:
By using client-specific batch normalization parameters, the underlying model can better generalize to new, unseen target domains.
The dedicated parameters for each client domain help mitigate the adverse effects of client-side domain shift, ensuring that the model performs well across different target domains.
Additionally, the established normalization process provides a stable input distribution, which can lead to faster convergence and improved training stability, both desired benefits for the CD-FSL setup.

In the server model, the \emph{classification head} $g$ is realized by a fully connected linear layer that receives the output of the adaptation module and projects that output to a specific set of labels.
Once the essential components of the source model have been defined, the overall decision function can be expressed as $\mathcal{F}(\phi_S, \psi_S, \nu_S) = g(\nu_S) \circ \mathcal{A}(\psi_S) \circ f(\phi_S)$. The training objective for our server-side classification task is defined as
\begin{equation} \label{eq_srv_pretrain}
    \argmin_{\phi_S, \psi_S, \nu_S} \mathcal{L}_{CE}(\mathcal{F}(\phi_S, \psi_S, \nu_S;\tilde{x}_S), y_S)
\end{equation}
where $\mathcal{L}_{CE}$ is the cross entropy loss regularized by label smoothing~\cite{10.5555/3454287.3454709} to further encourage robust output features and $y_S$ the ground truth label associated with $\tilde{x}_S$.
All optimizations are done with stochastic gradient descent and momentum until convergence.

In essence, the purpose of server pre-training on the source domain data set $D_S$ is to enable the learning and refinement of a set of features that are both disciminatory and transferable, serving to mitigate the discrepancy between the source and target domains.
This results in the adapted parameters of the server model being used as initial parameter set for client devices joining the FL process.
Technically, the parameter transmission can be performed in three distinct ways:
\begin{itemize}
    \item \textbf{On Demand}. Upon initiating the FL process for the first time, the central server transmits the model parameter set to the client. However, this approach has the disadvantage of incurring high communication costs associated with the initial transfer of all parameters, despite offering a flexible solution.
    \item \textbf{Pre-Configured}. At the time of installation, the client is equipped with a pre-trained model parameter configuration, thus obviating the need to update the weights at the initial stage and reducing the volume of communication. However, this approach may result in the parameter set becoming outdated and obsolete.
    \item \textbf{Differential Sync}. In the hybrid option, clients are provided with a pre-configured model parameter set, and the server is tasked with monitoring any discrepancies in weight relative to that baseline configuration. When new clients are introduced into the FL cycle, they are rapidly and effectively synchronised by the transfer of the weight delta between the baseline model and the current global model.
\end{itemize}
Furthermore, low-end client devices are integrated into the FL process of the central server in a step-by-step manner, in accordance with the adaptation described in Section~\ref{client_adapt_sec}. 

\subsection{Client Adaptation}\label{client_adapt_sec}
For each client the respective client model is designed from scratch with all aforementioned limitations in mind.
\begin{figure}[ht]
\centering
	\includegraphics[width=\linewidth]{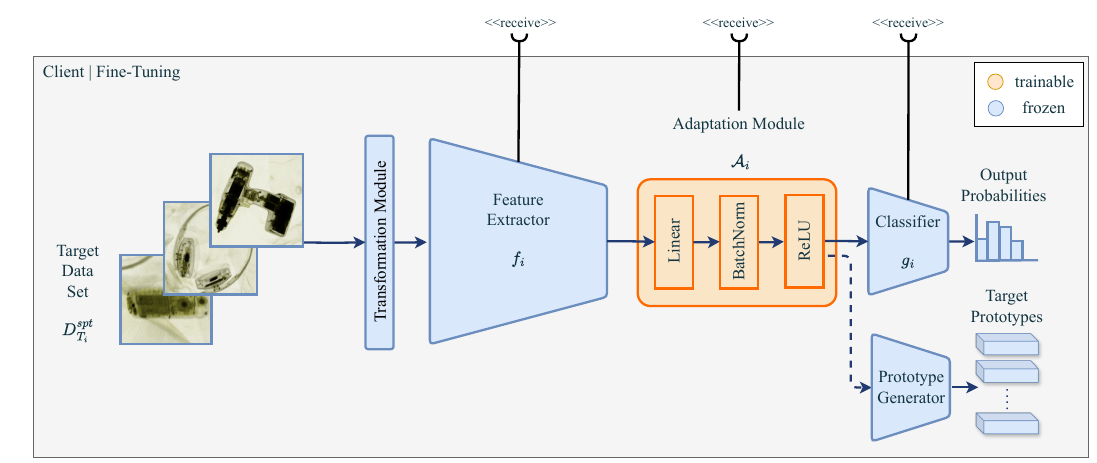}
	\caption{Model Architecture for client fine-tuning on reduced $k$-shot target support set $D^{spt}_{T_i}$. Module colors determine whether the parameters are frozen or trainable during training. Adapted from~\cite{roder2024orig}.
	  \label{fig:client_adapt}
    }
\end{figure}
As shown in Figure~\ref{fig:client_adapt}, the client model realizes the same training pipeline as the source model in order to ensure maximum parameter reuse.
After applying the pre-trained weights to the feature extractor $f_i$, the adaptation module $A_i$, and the classifier $g_i$, the parameter sets of the feature extractor and the classifier are frozen, resulting in these components being fixed during training.
This method is beneficial in many ways: disabling the backpropagation of error particularly through the deep feature extractor, thus avoiding computationally expensive gradient calculations of the corresponding weights, which in turn reduces the hardware requirements for the client model significantly.
Furthermore, the network training is optimized to achieve a convergent solution at an accelerated pace while maintaining stability.
Finally, the concept of keeping domain-specific information in a single, dedicated model component is supported by the exclusive fine-tuning of the parameters of the adaptation module.

The adaptation phase of client $i$ is carried out using a reduced $k$-shot target support set $D^{spt}_{T_i} \subset D_{T_i}$, where $k=\{3,5,10\}$ denotes the amount annotated samples per class, indicating scarcity of data in the target domain. 
We further evaluate and discuss the selection of $k$ in Section~\ref{experiments_sec}.
The training objective for the fine-tuning task of client $i$ is therefore defined as
\begin{equation} \label{eq_client_finetune}
    \argmin_{\psi_i} \mathcal{L}_{CE}(\mathcal{F}(\psi_i;\tilde{x}_l), y_l)
\end{equation}
where $\mathcal{F(\placeholderfunc)}$ is the client decision function parameterized with $\psi_i$, $\tilde{x}_l$ the transformed data sample drawn from the target domain data set $D^{spt}_{T_i}$, and the corresponding ground truth label $y_l$, respectively.

In the subsequent step (see Figure~\ref{fig:model_arch}), target prototypes are generated in a manner analogous to that employed for class prototypes in ProtoNet~\cite{snell2017prototypical} and FedProto~\cite{tan2021fedproto}, but enhanced with a particular fine-tuning strategy for \textit{FedAcross+}.
The selection of the prototypical representation is predicated on its computational simplicity, memory efficiency, and high interpretability.
Accordingly, the prototype representing the $n$-th class on client $i$ 
as mean vector of the embedded support points is denoted as:
\begin{equation} \label{eq_proto_create}
    \omega^{(n)}_{i} = \frac{1}{|D^{spt}_{T_i}, n|} \sum\limits_{(x_l, y_l) \in D^{spt}_{T_i}} \tau_i(\tilde{x}_l)
\end{equation}
where $\tau_i(\placeholderfunc)$ specifies the embedding function over the client-side feature extractor and adaptation module with $\tau_i(\tilde{x}_l) = \mathcal{A}_i(f_i(\tilde{x}_l))$.
The output set of the client adaptation is a collection of prototypes that have been adapted to align with the characteristics of the target data set.
\subsection{Client Inference}\label{client_infer_sec}
\begin{figure}[ht]
\centering
	\includegraphics[width=\linewidth]{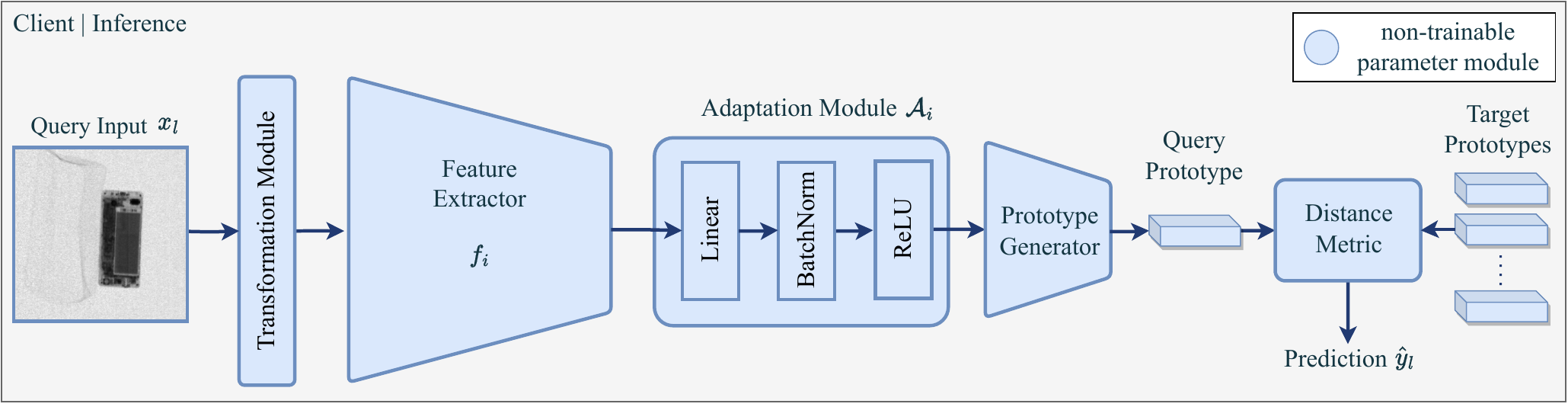}
	\caption{Client Inference. Source:~\cite{roder2024orig}.
	  \label{fig:model_client_infer}
    }
\end{figure}

Inference on the federated client device is a straightforward and resource-efficient process:
The end-to-end embedding pipeline, including feature extractor $f_i$ and adaptation module $\mathcal{A}_i$, is reused to project the unlabeled, transformed sample $\tilde{x}_l$, observed on client device $i$, in order to generate the corresponding query prototype as illustrated in Figure~\ref{fig:model_client_infer}\footnote{We show one of the X-ray images from the WeSort.AI waste detection scenario. The rechargeable battery of a cell phone has to be detected.}.
The embedded query vector is then injected into the \emph{Distance Metric Module} for computation of the pairwise L2 distance between the pre-computed target prototypes and the query vector.
The design choice to employ Euclidean distance as the distance metric in our proposed method is based on its integration with ProtoNet.
ProtoNet typically uses Euclidean distance due to several theoretical and practical reasons:
Theoretically, Euclidean distance is a member of Bregman divergences, which is significant because, in clustering problems, the centroid that minimizes the total distance to all points in a cluster is the mean of those points when using a Bregman divergence like Euclidean distance.
This aligns well with our method`s reliance on class prototypes, making Euclidean distance a natural fit for the client-side downstream classification tasks.
Practically, Euclidean distance is straightforward to compute and interpret, measuring the straight-line distance between points, which fits well with the concept of prototypes as central points in the embedding space~\cite{snell2017prototypical}.
Additionally, empirical studies have shown that ProtoNet performs better with Euclidean distance compared to cosine similarity, as it effectively captures the spread and central tendency of data points in the embedding space~\cite{https://doi.org/10.48550/arxiv.2110.05076}.
This is crucial for the prototype-based classification utilized in the inference step.
Consequently, the query sample is assigned to the class belonging to the nearest target prototype denoted as
\begin{equation} \label{eq_proto_inference}
    \hat{y}_l = \argmin_n \norm{\tau_i(\tilde{x}_l) - \omega^{(n)}_{i}}_2
\end{equation}
where $\hat{y}_l$ is the predicted label of sample $\tilde{x}_l$ observed on client $i$.
\subsection{Client Security and Prototype Upstreaming}\label{client_share_sec}
Federated client devices are not only able to benefit from the FL cycle, but are also able to contribute to it by sharing their locally refined knowledge without risking the direct exposure of sensitive information.
We argue that access to raw client data points is restraint in three ways:
\begin{enumerate}
    \item In contrast to the conventional FL approach, our methodology does not entail the interchanging of model gradients, thereby circumventing the potential issue of input data reconstruction from intercepted model gradients.
    \item Target prototypes are anchored at the mean of their respective class in the embedding space, restricted to only leak information in the same way that mean value statistics leak information~\cite{brinkrolfDifferentialPrivacyLearning2019}.
    \item It can be reasonably assumed that even in the event of an adversary successfully reconstructing the feature vector of a single data point and gaining access to the fine-tuned client model, the process of matching a raw client data point encoded by a deep backbone is considered to be a practically challenging and resource-intensive task.
\end{enumerate}
In general, reconstruction attacks require significant computational resources and the efficacy of reconstruction attempts can vary considerably depending on the specific underlying model, observed data set, and implemented security measures applied on client devices like \emph{Differential Privacy} for FL~\cite{wei2020}.

\vspace{0.5cm}
The \emph{Prototype Upstreaming} functionality enables client devices to send their generated target prototypes and adaptation module parameters back to the server, reducing the amount of data transferred while addressing bandwidth constraints and transmission latency.
The prototypes represent a highly compact injective encoding of the previously accumulated training data.
Similarly to how clients calculate target prototypes, the server can generate source prototypes by applying Equation~\eqref{eq_proto_create} on the source data set $D_S$ after pre-training the source model.
In comparison to target prototypes, these prototypes are more robust to outliers, yet they are also more specialized to the source data set.
To address this limitation, a range of techniques can be employed to enhance source prototypes with refined prototypes derived from client devices.
One such approach is the application of an optimal fusion strategy, as detailed in~\cite{tan2021fedproto}.
In order to facilitate the accelerated integration of new clients into the FL cycle, the initial weights of the adaptation module $\mathcal{A}$ are optimized with refined and processed parameters received from previous weight-sharing clients.
FedAvg~\cite{pmlr-v54-mcmahan17a} is one of the most well-known approaches to combine model parameters within the FL context, where weights are collected from remote devices and averaged on a central hub. This approach can be flawlessly integrated into our client-server setup.

\subsection{Client Data Stream Sampling Extension}\label{client_sample_sec}
\begin{figure}[h!]
\centering
	\includegraphics[width=\linewidth]{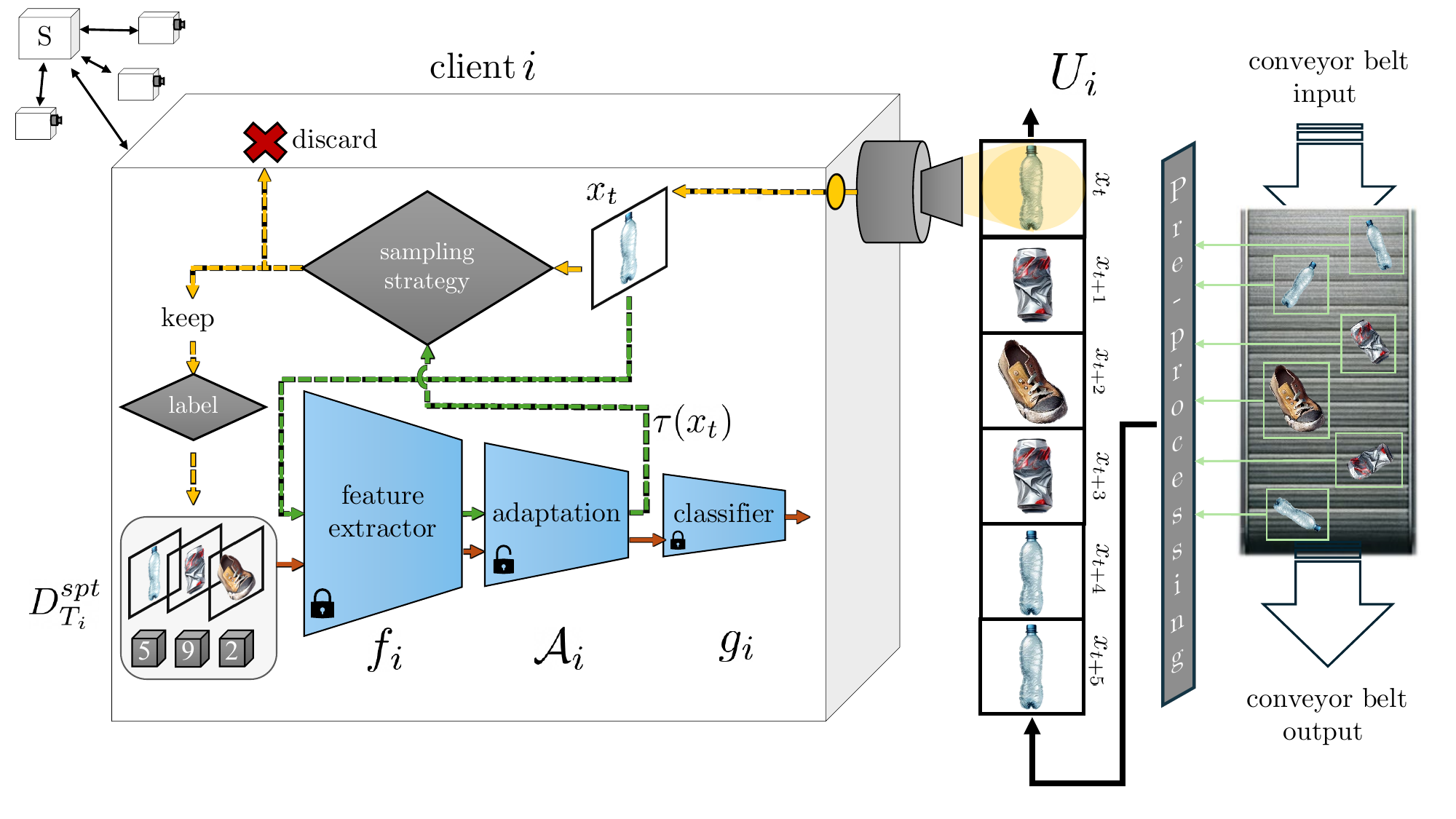}
	\caption{Federated Stream Sampling Extension. The embedding pipeline is colored in green, the sampling flow is colored in yellow and the training process is colored in red. Adapted from\cite{roeder2024sparse}.
	  \label{fig:stream_ext}
    }
\end{figure}

We propose an extension of baseline FedAcross by integrating a stream sampling strategy into the existing client adaptation and data collection workflow.
This is a modular compliment to the existing components, allowing the framework to be used in dynamic environments whilst maintaining the overall model architecture presented in Section~\ref{client_adapt_sec}.
In order to maintain focus on the topic of waste separation, we will initially examine an extended application scenario as described in Figure ~\ref{fig:stream_ext}.
We consider a sorting facility where waste items are processed on a conveyor belt.
A pre-trained FL client is deployed and tasked to perform accurate predictions on oncoming objects and improve the overall prediction accuracy over time by sampling specific items from the data stream used for client model fine-tuning.
For the sake of simplicity, in this section we reduce the scope to the consideration of a single client participating in a FL system.
Formally, item $x_t$ is observed from data stream $U_i$ at time $t$.
These waste items undergo a preprocessing stage where they are identified and framed to ensure they are adequately prepared for subsequent processing. This preprocessing step is essential for normalizing the input data and ensuring consistency across the data set.
Once the waste items are preprocessed, they are transmitted to the client instance, where a sophisticated sampling strategy is employed.
This strategy is required to be designed to evaluate each item based on its potential contribution to the $k$-shot support set $D^{spt}_{T_i}$:
The core of the sampling strategy involves a decision-making process that determines whether an item should be kept or discarded.
Each item is assessed for its informativeness and relevance to the learning task. This evaluation involves determining the item's representativeness and potential to improve the model's performance.
Items that do not meet the criteria for inclusion in the support set are discarded. This ensures that only the most relevant and informative samples are retained.
Conversely, items that are deemed valuable are kept for further processing.
Recently, a resource-aware sampling mechanism that satisfies our requirements derived from FL contraints has been proposed in the form of Volume Sampling for \emph{Streaming Active Learning} (VeSSal)~\cite{saran2023}.
This mechanism deploys labelling decisions based on the output of the penultimate layer while enforcing batch diversity, rendering it a valid approach to adequately populate $D^{spt}_{T}$.
Therefore, casting the yield of the embedding function $\tau(\placeholderfunc)$ as the penultimate layer output of the prediction model residing on the federated client, the currently observed sample $x_t$ is transformed using $\tilde{x}_t \gets \mathcal{T}(x_t)$ and set to be selected for labelling with probability $p_t$, proportional to its determinantal contribution defined as
\begin{equation} \label{eq_prob}
    p_t = \frac{q_t \cdot \tau(\tilde{x}_t)^T \hat{\Sigma}_t^{-1} \tau(\tilde{x}_t)}{\text{tr}\left( \frac{1}{t} \hat{\Sigma}_t^{-1} \sum_{j=1}^{t} \tau(\tilde{x}_t)\tau(\tilde{x}_t)^T \right)}
\end{equation} 
where $\hat{\Sigma}_t^{-1}$ denotes the inverse of the system covariance matrix over samples already selected for labelling and $q_t$ refers to the adaptive labelling frequency determined for the current stream observation.
For further details,~\cite{roeder2024sparse} offers an efficient sampling strategy tailored for federated clients.
The authors modified the sampling strategy from~\cite{saran2023} to better accommodate the requirements of federated learning and streaming data, thereby enhancing the algorithm's stability and immediate responsiveness across a substantial number of streaming observations.

For the items that are retained, the next critical step is labelling.
Each selected item is labeled with its corresponding category.
Labelling is a vital step as it provides the ground truth necessary for supervised learning.
Accurate labelling ensures that the support set is reliable and useful for fine-tuning the client model.
The labeled items are then aggregated into $D^{spt}_{T_i}$.
It is constructed to ensure that each category is represented equally, with $k$ samples per category.
This balance is crucial for FSL-inspired approach like FedAcross, as it helps the model to learn effectively from limited data.
The sampling strategy focuses on selecting samples that are diverse and most informative, choosing examples that best represent the variability within each category.
This enhances the model's generalization capabilities. Despite its limited size, the support set is highly informative due to the strategic selection of samples.

\subsection{FedAcross+ Algorithm}\label{sec_algorithm}
The pseudo-code of \textit{FedAcross+} presented in Algorithm~\ref{alg:f+} (Adapted from~\cite{roder2024orig}) is structured into three primary parts: Server Model Training, Client Adaptation, and Client Inference.
This section delineates each component in a detailed manner.
\begin{algorithm}[ht!]
\caption{FedAcross+ }\label{alg:f+}
\begin{flushleft}
\textbf{Input:} $D_S, \phi_S, \psi_S, \nu_S$ \;\;
$U_i$, $D^{spt}_{T_i}, \psi_i, i = 1, ..., I$\\
\end{flushleft}
\begin{algorithmic}[1]
   \phase{Server Model Training}
    \State Initialize server model.
    \ForAll{pre-train epoch}
        \ForAll{batch $(x_S, y_S) \in D_S$}
        \State Transform input $x_S$ by $\mathcal{T}$ 
        and compute loss by Eq. \ref{eq_srv_pretrain}.
        \State Update sever model parameters according to the loss.
        \EndFor
    \EndFor
    \setcounter{ALG@line}{0}
    \phase{Client Adaptation}
     \ForAll{new client $i$}
        \State Initialize client model with $\phi_i = \phi_S, \psi_i = \psi_S, \nu_i = \nu_S $.
        \State Freeze parameters $\phi_i$ and $\nu_i$.
        \If{SAMPLING\_ENABLED}
        \State Select data points from stream $U_i$ with probability $p_t$ according to Eq.~\eqref{eq_prob}. 
        \State Obtain labels for selected data points and assign to $D^{spt}_{T_i}$
        \EndIf
        \ForAll{local epoch}
             \ForAll{batch $(x_l, y_l) \in D^{spt}_{T_i}$}
                \State Transform input $x_l$ by $\mathcal{T}$
                and compute loss by Eq.~\eqref{eq_client_finetune}.
                \State Update parameters $\psi_i$ according to the loss.
             \EndFor
        \EndFor
        \State Create client prototypes by Eq.~\eqref{eq_proto_create}.
        \If{UPSTREAM}
        \State Upload client prototypes $\omega_{i}$ and parameters $\psi_i$.
        \EndIf
     \EndFor
   \setcounter{ALG@line}{0}  
   \phase{Client Inference}
    \State Compute label $\hat{y}_l$ by prototype inference Eq. \ref{eq_proto_inference}.
\end{algorithmic}
\end{algorithm}

In the first part, \emph{Server Model Training}, the process commences with the initialization of the server model (\texttt{line 1}).
Subsequently, the server model undergoes the pre-training phase (\texttt{line 2}).
During this phase, for every pre-training epoch, the algorithm iterates through each batch \((x_S, y_S)\) within the data set \(D_S\) (\texttt{line 3}).
For each batch, the input \(x_S\) is transformed utilizing the transformation function \(\mathcal{T}\), and the loss is computed in accordance with Equation~\eqref{eq_srv_pretrain} (\texttt{line 4}).
Following this, the server model parameters are updated based on the computed loss (\texttt{line 5}).
This iterative process continues until the completion of all pre-training epochs (\texttt{line 6}).

The second part, \emph{Client Adaptation}, addresses the initialization and adaptation of new client models (\texttt{line 1}).
For each new client \(i\), the client model is initialized with parameters \(\phi_i = \phi_S\), \(\psi_i = \psi_S\), and \(\nu_i = \nu_S\) (\texttt{line 2}).
Subsequently, the parameters \(\phi_i\) and \(\nu_i\) are frozen (\texttt{line 3}).
If data stream sampling is enabled, as indicated by the \texttt{SAMPLING\_ENABLED} flag, the algorithm proceeds to sample data points from the stream \(U_i\) (\texttt{line 5}).
These sampled data points are then labeled and assigned to the data set \(D_{T_i}^{spt}\) (\texttt{line 6}).
During the local training phase on the client, the algorithm iterates through all local epochs (\texttt{line 8}).
Within each epoch, it processes each batch \((x_l, y_l)\) in the data set \(D_{T_i}^{spt}\) (\texttt{line 9}).
The input \(x_t\) is transformed using the transformation function \(\mathcal{T}\), and the loss is computed following Equation~\eqref{eq_client_finetune} (\texttt{line 10}).
The parameters \(\psi_i\) are subsequently updated according to the computed loss (\texttt{line 11}).
Upon completion of the local epochs, client prototypes are created utilizing Equation~\eqref{eq_proto_create} (\texttt{line 14}).
If the \texttt{UPSTREAM} condition is satisfied, the client prototypes \(\omega_i\) and parameters \(\psi_i\) are upstreamed to the server $S$ (\texttt{line 16}).

The final part, \emph{Client Inference}, involves the computation of the label \(\hat{y}_i\) via prototype inference using Equation~\eqref{eq_proto_inference} (\texttt{line 1}). This step ensures that the client is capable of inferring labels based on the trained model and generated target prototypes.
The comprehensive framework of \textit{FedAcross+} facilitates efficient server and client model training, adaptation, and inference within a federated learning environment.

\section{Experiments}\label{experiments_sec}
In this section, we explain the experiments of our approach replicating a garbage classification scenario\footnote{The original x-ray and multi-spectral image data could not be made available for copyright reasons, but the data in the experiments are sufficiently similar.}. 
For the sake of simplicity, it is assumed that the data provisioning for both the FL server and the FL client is static.
The efficacy of stream sampling strategies has already been thoroughly evaluated in~\cite{saran2023} and subsequently applied and tested in the context of FL in~\cite{roeder2024sparse}.
\subsection{Implementation Setup}
The experiments are conducted with the objective of reflecting the real-world challenges that arise due to domain shift when analyzing observations, the scarcity of annotations on target samples and the additional constraints imposed by the FL environment.
For each experiment, the feature extractor of the server instance is first initialized with a ResNet-34 or ResNet-50 architecture using corresponding weights derived from pre-training on ImageNet~\cite{ILSVRC15}, respectively.
Furthermore, the weight parameters of the adaptation module and the classifier's linear layer are initialized with values drawn from a normal distribution. 
Conversely, the weights of the batch normalization layer are initialised through the utilisation of the Xavier normal initialisation method~\cite{pmlr-v9-glorot10a}.
The primary objective of the \emph{server pre-training} is the optimization of the server model to recognize source domain-specific classes by minimizing the cross-entropy loss from Equation~\eqref{eq_srv_pretrain} on the full source data set $D_S$.
Following~\cite{guo2020broader} for a pre-training configuration in conjunction with few-shot learning downstream tasks, a mini-batch stochastic gradient descent optimizer with initial learning rate of 0.01, momentum of 0.9 and weight decay of 0.001 is deployed.
The learning rate is gradually diminished through the implementation of a learning rate scheduling strategy.
Server training runs 300 epochs, processing randomly shuffled mini-batches of size 128 per epoch, with optional early stopping on model convergence. The server-side transformation module follows the recommendation from~\cite{kim2022finetune} by applying horizontal flipping, random resized cropping and color jittering to augment and alter the input data.
Following the pre-training step, the Flower-based server establishes a gRPC\footnote{Remote Procedure Call protocol developed by Google.} network connection and awaits the arrival of clients to participate in the FL round. 

The \emph{Fine-tuning} phase starts by booting up client instances announcing their availability to the server with a pre-configured parameter set transferred from the pre-training stage.
Furthermore, to replicate the scarcity of labeled target data, each client $i$ is only permitted to access its respective $k$-shot support set $D^{spt}_{T_i}$ during training, where each support set is randomly drawn from the corresponding subdomain of the DA data set under inspection.
The objective of the client training is to fine-tune the client model by optimizing the cross-entropy loss as defined in Equation~\eqref{eq_client_finetune}.
The training setup for clients and server is identical, except that the client has a learning rate set to 0.1 and a mini-batch size of 32.
For the purposes of simulation and fair comparison, the number of training epochs is set to 200 for a single federated round and clients can access all support samples and their corresponding labels immediately.
In real world scenarios, the training epochs would be split and distributed over the number of federated rounds.
Once the fine-tuning process is complete, the client generates the target prototypes based on Equation~\eqref{eq_proto_create}.
The test accuracy is reported by assessing each client individually using mean-centroid classification on its respective hold-back test set and computing the average classification accuracy over five runs.

\subsection{Model Evaluation}
In order to provide an adequate benchmark for our model, we initially evaluate our method against approaches of source-free unsupervised DA as set up in~\cite{zhang2022lccs}, focusing on single-domain performance.
We chose the official Office-31~\cite{10.1007/978-3-642-15561-1_16} and OfficeHome~\cite{venkateswara2017deep} benchmark data sets to be the most suitable fit for evaluation purposes, as the contained subdomains are based on images of real-world objects with visual differences in terms of backgrounds, lighting conditions and viewpoints.
A total of 31 object classes with 4110 images are present in the Office-31 data set scattered over three domains: Amazon (\textbf{A}), DSLR (\textbf{D}) and Webcam (\textbf{W}).
The OfficeHome data set used in the second experiment contains 15500 images and 65 object classes divided into four domains: Art (\textbf{A}), Clipart (\textbf{C}), Product (\textbf{P}) and Real World (\textbf{RW}).
For both experiments, we initially select a source domain (e.g. $D_S = \mathbf{A}$ selects domain \textbf{A} as source), pre-train the server model on it and clone the model to fine-tune it on the remaining domains, with $\mathbf{A} \rightarrow \mathbf{W}$ and $\mathbf{A} \rightarrow \mathbf{D}$ illustrating the mean-centroid classification task on the test set of the target domains $\mathbf{W}$ and $\mathbf{D}$, respectively.
The performance of \textit{FedAcross+} is evaluated using a ResNet-50 feature extractor, as all competing methods use the latter.
First, we compare our approach with source-free DA methods that allow access to the \emph{full target} data set for fine-tuning: SHOT~\cite{10.5555/3524938.3525498}, SFDA~\cite{9528982} and SDAA~\cite{DBLP:journals/corr/abs-2102-09003}.
We also compare our approach to recent state-of-the-art few-shot adaptation methods FLUTE~\cite{triantafillou2021learning}, which develops a universal template based on multiple source data sets, and LCCS~\cite{zhang2022lccs}, which adapts batch normalization statistics to target samples.

\begin{table}[ht]
\centering
\caption{Results\tablefootnote{Results referenced from~\cite{zhang2022lccs}; we use \textit{FedAcross(+)} in the table  to indicate interchangeability of \textit{FedAcross}~\cite{roder2024orig} and \textit{FedAcross+} in terms of experimental results.} with ResNet-50 baseline, centralized source-free DA and few-shot transfer learning methods on Office-31. "$\rightarrow$" indicates a domain change, $k$ denotes the number of labeled samples per class available fine-tuning. Adapted from~\cite{roder2024orig}.}
\label{exp_1_tab}
\begin{tabular}{cccccccccc}
\toprule
 \multirow[t]{4}{*}{ \textbf{Method} } & \multirow[t]{2}{*}{$\textbf{k}$} & \multicolumn{7}{c}{ \textbf{Office-31} } \\
 \hline
 \multirow[t]{4}{*}{ } & \multirow[t]{2}{*}{ } & \multicolumn{2}{c}{ \textbf{ $D_S = \mathbf{A}$ } } & \multicolumn{2}{c}{ \textbf{$D_S = \mathbf{W}$ } } & \multicolumn{2}{c}{ \textbf{$D_S = \mathbf{D}$ } } & \multicolumn{1}{c}{ } \\
 & & $\mathbf{A} \rightarrow \mathbf{W}$ & $\mathbf{A} \rightarrow \mathbf{D}$ & $\mathbf{W} \rightarrow \mathbf{A}$ & $\mathbf{W} \rightarrow \mathbf{D}$ & $\mathbf{D} \rightarrow \mathbf{A}$ & $\mathbf{D} \rightarrow \mathbf{W}$ & $\mathbf{Avg}$ \\
 \hline
 Baseline & - & 68.4 & 68.9 & 60.7 & 99.3 & 62.5 & 96.7 & 76.1 \\
 SHOT & all & 90.1 & 94.0 & \underline{74.3} & \textbf{99.9} & \underline{74.7} & 98.4 & 88.6 \\
 SFDA & all & 91.1 & 92.2 & 71.2 & 99.5 & 71.0 & 98.2 & 87.2 \\
 SDDA & all & 82.5 & 85.3 & 67.7 & 99.8 & 66.4 & \textbf{99.0} & 83.5 \\
 FLUTE$^*$ & 5 & 84.6 & 88.2 & 66.4 & 99.1 & 66.4 & 95.3 & 83.3 \\
 $\mathrm{LCCS}^*$ & 5 & 92.8 & 91.8 & \textbf{75.1} & \underline{99.9} & \textbf{75.4} & \underline{98.5} & \underline{88.9} \\
 \hline
 \textit{FedAcross(+)} & 5 & 89.4 & 90.4 & 63.5 & 94.5 & 60.0 & 90.4 & 81.4\\
 \textit{FedAcross(+)} & 10 & \textbf{97.4} & \textbf{98.5} & 71.1 & 98.6 & 71.0 & 97.4 & \textbf{89.0}\\
 \bottomrule
\end{tabular}
\end{table}

The experimental outcome on Office31 (Table~\ref{exp_1_tab}) shows that \textit{FedAcross+} delivers on par adaptation results against all competitors despite the more difficult conditions caused by the FL setup:
Although LCCS yields the best overall adaptation performance (88.9\%) with five labeled samples per class, \textit{FedAcross+} produces the best overall adaptation results of all methods under inspection with $k = 10$ (89.0\%).
In conclusion, our approach presents a more practical solution than LCCS for cross-device FL scenarios:
First, the \textit{FedAcross+} approach exhibits increased flexibility as client adaptation does not depend on the number of batch normalization layers of the feature extractor, rendering it more versatile and applicable to a wider range of network architectures.
Second, unlike \textit{FedAcross+}, the LCCS method requires a two-stage adaptation process with a computationally intensive grid search in the first stage.
This demanding computational task is required to approximate the optimal parameter configuration for its learnable coefficients, which are then applied to kick-start the gradient update stage.


The results on the more challenging OfficeHome benchmark data set, as presented in Table~\ref{exp_3_tab}, demonstrate that the unsupervised DA method SHOT outperforms all other competitors in that scenario, highlighting the difficulty of adaptation in low data regimes (71.8\% SHOT - 70.9\% \textit{FedAcross+}, $k=10$).
We argue against SHOT that it requires on average six times the amount of (unlabeled) data samples per class in the OfficeHome setup (59.6 images/class for SHOT - $k$ images/class for \textit{FedAcross+}, $k = 10$) to achieve only marginally superior overall accuracy than \textit{FedAcross+}.

Two further insights emerge from the results regarding the issue under consideration: There is a trade-off between the number of parameters that need to be transferred to the client initially (communication efforts) and the on-client adaptation and prediction performance determined by the selection of the feature extractor.
Furthermore, the number of ground truth annotations $k$, is a crucial factor in enhancing the prediction accuracy according to the specific needs of client operators.

\begin{table}[ht]
\caption{Results\tablefootnote{Results referenced from~\cite{zhang2022lccs}; we use \textit{FedAcross(+)} in the table  to indicate interchangeability of \textit{FedAcross}~\cite{roder2024orig} and \textit{FedAcross+} in terms of experimental results.} with ResNet-50 baseline, centralized source-free DA and few-shot transfer learning methods on OfficeHome. "$\rightarrow$" indicates a domain change, $k$ denotes the number of labeled samples per class available fine-tuning. Adapted from~\cite{roder2024orig}.}
\label{exp_3_tab}\resizebox{\linewidth}{!}{
\begin{tabular}{cccccccccccccccc}
\toprule
 \multirow[t]{4}{*}{ \textbf{Method} } & \multirow[t]{2}{*}{$\textbf{k}$} & \multicolumn{13}{c}{ \textbf{OfficeHome} } \\
 \hline
 \multirow[t]{4}{*}{ } & \multirow[t]{2}{*}{ } & \multicolumn{3}{c}{ \textbf{ $D_S = \mathbf{A}$ } } & \multicolumn{3}{c}{ \textbf{$D_S = \mathbf{C}$ } } & \multicolumn{3}{c}{ \textbf{$D_S = \mathbf{P}$ } } & \multicolumn{3}{c}{ \textbf{$D_S = \mathbf{RW}$ } } & \multicolumn{1}{c}{ } \\
 & & $\mathbf{A} \rightarrow \mathbf{C}$ & $\mathbf{A} \rightarrow \mathbf{P}$ & $\mathbf{A} \rightarrow \mathbf{RW}$ & $\mathbf{C} \rightarrow \mathbf{A}$ & $\mathbf{C} \rightarrow \mathbf{P}$ & $\mathbf{C} \rightarrow \mathbf{RW}$ & $\mathbf{P} \rightarrow \mathbf{A}$ & $\mathbf{P} \rightarrow \mathbf{C}$ & $\mathbf{P} \rightarrow \mathbf{RW}$ & $\mathbf{RW} \rightarrow \mathbf{A}$ & $\mathbf{RW} \rightarrow \mathbf{C}$ & $\mathbf{RW} \rightarrow \mathbf{P}$ & $\mathbf{Avg}$ \\
 \hline
 Baseline & - & 34.9 & 50.0 & 58.0 & 37.4 & 41.9 & 46.2 & 38.5 & 31.2 & 60.4 & 53.9 & 41.2 & 59.9 & 46.1 \\
 SHOT & all & \underline{57.1} & \textbf{78.1} & \textbf{81.5} & \underline{68.0} & \textbf{78.2} & \textbf{78.1} & \underline{67.4} & \underline{54.9} & 82.2 & 73.3 & 58.8 & \textbf{84.3} & \textbf{71.8} \\
 SFDA & all & 48.4 & 73.4 & 76.9 & 64.3 & 69.8 & 71.7 & 62.7 & 45.3 & 76.6 & 69.8 & 50.5 & 79.0 & 65.7 \\
 FLUTE$^*$ & 5 & 49.0 & 70.1 & 68.2 & 53.8 & 69.3 & 65.1 & 53.2 & 46.8 & 70.8 & \textbf{59.4} & 51.7 & 77.3 & 61.2 \\
 $\mathrm{LCCS}^*$ & 5 & \textbf{57.6} & 74.5 & \underline{77.0} & 60.0 & 71.5 & 70.9 & 59.2 & 54.7 & 75.9 & 69.2 & \textbf{61.2} & 81.5 & 67.8 \\
 \hline
 \textit{FedAcross(+)} & 5 & 45.9 & 68.9 & 66.1 & 53.9 & 67.6 & 64.8 & 55.8 & 47.2 & 67.2 & 59.5 & 48.1 & 73.3 & 59.9 \\
 \textit{FedAcross(+)} & 10 & 56.7 & \underline{77.0} & 76.3 & \textbf{69.1} & \underline{76.7} & \underline{74.6} & \textbf{69.5} & 59.4 & \underline{76.6} & \underline{72.4} & \underline{60.4} & \underline{81.5} & \underline{70.9} \\
 \bottomrule
\end{tabular}}
\end{table}

In order to assess the efficacy of our methodology with regards to the \emph{classification of waste items}, the DA benchmark data sets OfficeHome and DomainNet~\cite{peng2019moment} (30 waste object classes, Clipart and Real domain) are modified to include only those items that are typically observed in waste sorting scenarios\footnote{Waste object classes are specified in the \textit{FedAcross} sources.}.
In the experimental setup, the waste sorting service provider ($\mathbf{Srv}$) pre-trains its source model on all available waste object classes.
Subsequently, the waste sorting facilities ($\mathbf{Cl}$) fine-tune their local model on a specialized, randomly sampled subset of ten classes, respectively.
Table~\ref{exp_2_tab} presents the prediction accuracy averaged over five runs with $k$ = \{0, 3, 5, 10\}.
The results for OfficeHome (Waste) demonstrate that \textit{FedAcross+} effectively improves the prediction accuracy on federated client devices with $k>3$ in a photorealistic adaptation task, scaling with an increase in the number of annotated data points.
The enhanced performance observed in the more challenging DomainNet (Waste) adaptation tasks across two domains with a larger distributional gap serves to underscore the flexibility of the \textit{FedAcross+} approach.
\begin{table}[ht]
\centering
\caption{Adaptability of \textit{FedAcross+} in a waste sorting scenario. "$\rightarrow$" indicates a domain change, $k$ denotes the number of labeled samples per class available fine-tuning. Source:~\cite{roder2024orig}.}
\label{exp_2_tab}
\begin{tabular}{cccccc}
\toprule
 \multirow[t]{2}{*}{} & \multicolumn{2}{c}{OfficeHome (Waste)} & \multicolumn{2}{c}{DomainNet (Waste)}\\
 \textbf{k} & $Srv_{\mathbf{RW}} \rightarrow Cl_{\mathbf{P}}$ & $Srv_{\mathbf{P}} \rightarrow Cl_{\mathbf{RW}}$ & $Srv_{\mathbf{C}} \rightarrow Cl_{\mathbf{R}}$ & $Srv_{\mathbf{R}} \rightarrow Cl_{\mathbf{C}}$ \\
 \hline
 \textbf{0} & 87.82$\pm$0.26 & 78.68$\pm$0.86 & 54.51$\pm$0.12 & $65.0\pm$0.50\\
 \textbf{3} & 84.42$\pm$0.25 & 75.73$\pm$0.73 & 54.74$\pm$0.06 & 69.0$\pm$0.65 \\
 \textbf{5} & 89.43$\pm$0.45 & 84.44$\pm$0.29 & 57.77$\pm$0.24 & 76.87$\pm$0.25\\
 \textbf{10} & 93.45$\pm$0.56 & 88.91$\pm$0.19 & 66.48$\pm$0.32 & 83.18$\pm$0.53\\
 \bottomrule
\end{tabular}
\end{table}

\begin{figure}[h!t]
\centering
	\includegraphics[width=\linewidth]{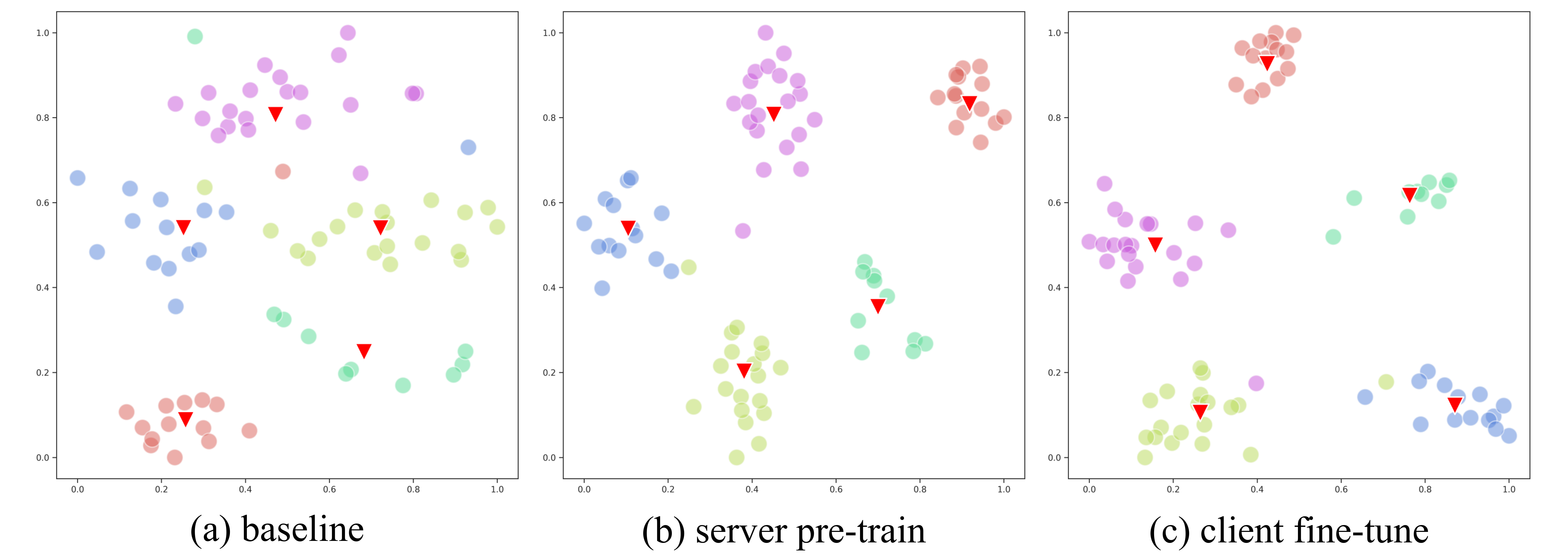}
	\caption{t-SNE plot of target class data (color-coded dots) and respective target class prototype (red triangles). Source:~\cite{roder2024orig}.
	  \label{fig:tsne}
    }
\end{figure}

Finally, we utilize the interpretable nature of our approach to visualize the separation progress over multiple adaptation stages using t-SNE~\cite{7b54165e73a3424b8820136bcf61ca89} on the Office-31 classification task $\mathbf{A} \rightarrow \mathbf{W}$.
One of the key objectives of our approach is to identify the optimal projection that will facilitate the closer grouping of samples belonging to the same class while simultaneously creating a greater separation between samples from different classes. This results in the generation of prototypes that are more representative of the underlying data.
In Figure~\ref{fig:tsne}, the plots illustrate the evolution of target sample feature projections of five classes using: \emph{(a)} an off-the-shelf ResNet-50 backbone, \emph{(b)} a model pre-trained on $D_S$ and \emph{(c)} a model pre-trained on $D_S$ and fine-tuned on $D^{spt}_{T_i}$ with their prototypes represented as red rectangles, respectively.
Although baseline prototypes are relatively close to each other, the pre-training facilitates the initial separation of samples into distinct classes.
Fine-tuning further reduces the similarity between samples of the same class, providing a suitable basis for the application of a nearest-centroid classifier.
\section{Conclusion}
In this work, we present \emph{FedAcross+}, a computationally efficient FL approach that offers a readily deployable solution for target adaptation tasks in the context of resource constraints and distributional shifts across disparate data silos.
This paper demonstrates the scalability and flexibility of our method by exemplifying an image recognition task motivated by intelligent waste sorting systems.
The employment of prototype-based few-shot learning in conjunction with cross-device domain adaptation techniques enables our model to achieve competitive results in a federated server-client environment, while simultaneously minimizing communication and computation efforts.
We further propose to extend the baseline framework by integrating a stream sampling pipeline into the client adaptation and data collection workflow, enabling its use in dynamic environments while maintaining the overall model architecture.
A comprehensive series of experiments conducted on both public and industry data sets have substantiated the viability of our presented approach in real-world production environments.

Although our current approach yields considerable insights, it also presents a number of avenues for \emph{future research}.
An expedient continuation of our research could involve strategically detecting and handling concept drifts that occurs on federated clients observing dynamically evolving data streams, a prospect that offers considerable potential for advancement.
This progression would significantly improve the model's adaptability and performance in real-world, dynamic scenarios.
Another crucial element of future work in this area will be the assessment of the quality of data points obtained from streaming data, with the objective of ensuring that the most informative samples contribute to the learning process.
Furthermore, investigating the influence of these advancements on privacy preservation and communication efficiency in federated settings could yield valuable insights, aligning with the growing demand for secure and scalable machine learning solutions.
Ultimately, these endeavours would contribute to the development of more sophisticated, efficient, and practical federated learning systems, capable of handling the complexities of real-world data distributions and applications.

\begin{credits}
\subsubsection{\ackname} This work was funded in part by the European Regional Development Fund (ERDF) under grant FKZ:2404-003-1.2 and in part by the ProPere THWS program.
M. R. thanks the Center for Artificial Intelligence (CAIRO) Würzburg for their support.
\end{credits}

\bibliographystyle{splncs04}
\bibliography{main}

\ \\
\end{document}